RESEARCH ARTICLE

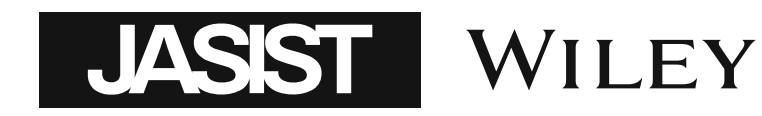

# Will sentiment analysis need subculture? A new data augmentation approach

**Zhenhua Wang**[1] | **Simin He**[2] | **Guang Xu**[1] | **Ming Ren**[1]

[1]School of Information Resource Management, Renmin University of China, Beijing, China

[2]Faculty of Humanities and Social Sciences, Beijing University of Technology, Beijing, China

**Correspondence**

Ming Ren, School of Information Resource Management, Renmin University of China, 401, Information Building, No. 59 Zhongguancun Street, Haidian District, Beijing, China.

Email: renm@ruc.edu.cn

**Funding information**

Key Projects of the National Social Science Foundation of China, Grant/Award Number: 21ATQ008

**Abstract**

Nowadays, the omnipresence of the Internet has fostered a subculture that congregates around the contemporary milieu. The subculture artfully articulates the intricacies of human feelings by ardently pursuing the allure of novelty, a fact that cannot be disregarded in the sentiment analysis. This paper aims to enrich data through the lens of subculture, to address the insufficient training data faced by sentiment analysis. To this end, a new approach of subculture-based data augmentation (SCDA) is proposed, which engenders enhanced texts for each training text by leveraging the creation of specific subcultural expression generators. The extensive experiments attest to the effectiveness and potential of SCDA. The results also shed light on the phenomenon that disparate subcultural expressions elicit varying degrees of sentiment stimulation. Moreover, an intriguing conjecture arises, suggesting the linear reversibility of certain subcultural expressions.

## 1 | INTRODUCTION

Sentiment analysis plays a crucial role in identifying, extracting, and categorizing the emotions or opinions expressed in text form of human communication (Medhat et al., 2014; Zhang et al., 2018). It aims to provide insights into people's feelings, thoughts, behaviors, attitudes, and perceptions towards various topics, products, or services, thereby to offer informed decision-making (Feldman, 2013; Hussein, 2018). For example, businesses can utilize sentiment analysis to gain a better understanding of customers' preferences, interests, and emotions towards their offerings, thus improving customer engagement, loyalty, satisfaction, and retention (Bueno et al., 2022; Kauffmann et al., 2020). Medical and healthcare institutions can monitor patients' mental states, thereby facilitating early intervention, diagnosis, and treatment (Gohil et al., 2018). The analysis of sentiments expressed in public opinion posts, comments and messages also empowers individuals and organizations to understand their reputation, manage financial affairs, track, and predict social events (Bi, 2022; Paltoglou, 2016; Ren et al., 2021; Sinha et al., 2022; Yildirim, 2022). Also, sentiment analysis can assist policymakers and government agencies in shaping political campaigns, strategies, and messaging, as well as developing assessments about public policy and resource allocation (Chung & Zeng, 2016; Verma, 2022). It can even have an impact on citation measurement, and so forth (Cruz et al., 2016; Liu et al., 2023; Melo et al., 2019; Yan et al., 2020).

Sentiment analysis algorithms reply heavily on abundant training data to effectively identify patterns. However, datasets collected from real-world scenarios often suffer from a shortage of data. The manual aggregation of domain-specific training datasets can be a labor-intensive and costly endeavor. This challenge underscores the critical need for an exploration into data augmentation strategies, which has secured a broad footprint in the realm of image processing, typically implemented through transformations (Alqudah et al., 2023; Niu et al., 2023; Shorten & Khoshgoftaar, 2019; Xiang et al., 2021).

With regard to natural language processing (NLP), instituting universal solutions proves exceedingly challenging. Instead, unique tasks dictate the introduction of bespoke enhancement methodologies. This could span the spectrum from commonsense reasoning and automatic translation to text comprehension and generation, extending to entity extraction and sentiment analysis (Abonizio et al., 2021; Hsu et al., 2021; Liesting et al., 2021; Liu et al., 2021; Shen et al., 2020; Xiang et al., 2021; Yang et al., 2020). Regarding sentiment analysis, popular augmentation strategies encompass synonym substitution (Zhang et al., 2015) and a technique known as easy data augmentation (EDA) (Wei & Zou, 2019). These methods are strategically utilized to fortify the training process and improve the model performance, highlighting the key role of data augmentation in this domain.

However, existing methods have fallen short in encapsulating the breadth and diversity of sentiments expressed in the present socio-cultural milieu, particularly by younger demographics. This group, as the driving force behind societal progress and cultural evolution, tends to employ more captivating and memorable sentiment expressions, especially on linguistically rich social media platforms (Bennett & Kahn-Harris, 2020). These arresting forms of expression have subtly given rise to a subculture. Their sentiment communication, deeply rooted in subcultural dynamics, is highly distinctive and impactful, demanding a greater attention in sentiment analysis. This cohort's innovative use of language, reflecting the zeitgeist and the nuances of contemporary youth culture, presents an untapped opportunity for a more contextual and inclusive sentiment analysis, thereby necessitating further exploration.

This study aims to integrate the subcultural expression within the framework of data augmentation, exploring its potential to enhance the efficacy of sentiment analysis models. A novel data augmentation approach termed SCDA is introduced, which specifically targets the multiform nature of contemporary social discourse. Its design is motivated by the following considerations.

1. In the current landscape of subculture, sentiment articulation is encapsulated in a diverse array of expressions and language patterns. This is evident in the trend of homophonic stems for example, which enjoy widespread popularity. Such trends are reflections of our present society, life, political milieu, and prevailing attitudes (Bennett & Kahn-Harris, 2020; Guerra, 2020; Jensen, 2018; Sun & Lee, 2020). Since the model's primary function is to aid human decision-making, it should ideally learn from daily habits and dialogues in subculture, thus attaining a superior understanding of contemporary sentiment currents.
2. The alteration of expression within the subculture does not tamper with the original sentiment, but instead modulates the intensity of sentiment, either amplifying or diminishing it (Bennett & Kahn-Harris, 2020). This modulation can be harnessed to create a more diverse data augmentation. The inventive rewording can serve as an additional dimension for data enhancement.

Undoubtedly, the realm of subcultures is constantly evolving, making it nearly impossible to fully encapsulate all aspects of subcultural expression. Therefore, this paper focuses on the aspects that have lasting relevance in today's globalized and interconnected society, marking the beginning of a new chapter in this interesting and meaningful discourse. Our decision to explore the phenomenon of spoonerism originates from its historical significance and enduring popularity (Bennett & Kahn-Harris, 2020). Reverend William Archibald Spooner's linguistic slip-ups have persisted through time, transforming into a beloved form of linguistic entertainment. This choice encapsulates the lasting impact of manipulating language, resonating across different eras. It serves as a method of self-expression by rearranging subjects, objects, or word combinations within the text (Yule, 2022). For example, transforming "John Watson, you discover blind spots" into "Blind spots, you discover John Watson" introduces an intriguing twist. Additionally, our exploration encompasses the homophone meme expression (Schmidt & de Kloet, 2017; Wong et al., 2021), the inversion rhetoric expression (Elton, 2016; Fitz & Chang, 2017), the emoji encryption expression (Franco & Fugate, 2020), the decomposed expression (Chen, 2021), the mobile data economizing expression (Zea & Heekyoung, 2019).

Formally, we deliberate upon the above prevalent facets of subculture. These creative expressions emulate original textual frameworks, amplifying the potency of words and phrases that might have been subjected to semantic dilution. Through the intriguing process of deconstruction and reconstruction, they spawn novel linguistic effects, fortifying the original meanings while concurrently intensifying the sentiment undercurrents embedded within. This ingenious manipulation of language acts as a magnet, captivating increased attention, and engagement. A series of corresponding generators is designed to generate enhanced text for each training data, to capture the expressions of texts within the subcultures. Consequently, each data embodies its own unique perspective from multi-distinct vantage points, thus resulting in manifold augmentation, providing the model with additional stimuli. Extensive experimentation attests to the effectiveness of SCDA.

The main contributions are as follows:

1. This paper delves into the fusion of subculture and sentiment analysis, presenting the idea of subculture-based data augmentation for enhancing sentiment analysis.
2. Specific generators that capture the expressions within subcultures are designed, and extensive experiments provide compelling evidence for the effectiveness.
3. An observation emerges that subcultural expressions elicit varying levels of stimulation for the model in different contexts. Also, there is a suggestion of a potential linear relationship between certain subcultural expressions.

## 2 | RELATED WORK

### 2.1 | Sentiment analysis data augmentation

With the exponential proliferation of user-generated text across the web, the use of sentiment analysis models becomes crucial. Sentiments inject a measure of subjectivity, serving as a vital component within human interactions. Consequently, sentiment analysis becomes instrumental in deciphering and comprehending this inherent subjectivity and its subsequent variations. Its potential is immense, notably as an integral part of emerging technologies. These innovations autonomously process vast volumes of data, subsequently extracting invaluable knowledge and insight from what would otherwise be a disarray of unstructured information.

Sentiment analysis necessitates voluminous amounts of training data for proficient the recognition of these patterns. In the development of such systems, however, real-world labeled datasets frequently encounter data scarcity issues. This dearth of data can compromise the operational efficiency of these models in pragmatic scenarios. The shortage of textual resources has also persistently posed a challenge in numerous NLP tasks. This predicament often jeopardizes the quality of samples and skews data distribution, infringing upon the foundational assumptions underpinning the majority of learning algorithms.

Despite the recency of text data augmentation methods in sentiment analysis, they present promising solutions to mitigate data scarcity (Abonizio et al., 2021). These ways execute class-preserving operations on the primary data source and primarily rely on strategies such as lexical substitution (Wei & Zou, 2019; Xiang et al., 2021), word embedding interpolation (Jin et al., 2023), and neural model generation (Gupta, 2019). The latter two ways fall under the umbrella of neural network modeling, which often induce significant overhead into the pipeline, amplifying the training duration. In addition, the situation where text retains its original label after being opaquely disturbed is more complex. Hence, ways premised on lexical substitution prove to be particularly valuable, such as EDA (Wei & Zou, 2019), a popular approach in this regard, which employs four distinct operations, i.e., synonym replacement, random insertion, swapping, and deletion.

However, given the evolution of contemporary society, the realms of politics, economy, and culture have shaped the dominant discourse groups on social media, prompting them to employ an array of expressive forms to articulate their sentiments (Bennett & Kahn-Harris, 2020). For instance, the negative sentiment within the phrase "little girl selling match" is frequently reinterpreted as "little match selling girl," a creative inversion designed to capture increased attention from digital spectators. These arresting forms of expression have subtly given rise to a subculture, gaining traction across various entertainment communities and online platforms. However, prior data augmentation strategies have largely overlooked these shifts in expression.

### 2.2 | Subcultural expression

Subculture encompasses communities sharing cultural practices, values and beliefs, and reflects the unique social, cultural, and political context of contemporary world. The accelerated transformation of society, the iterative updating of media technology, and the pervasive penetration of business logic have collectively provided an ideal environment for the incubation and cultivation of subculture (Bennett & Kahn-Harris, 2020). Subculture has become deeply embedded themselves as a way for the public to express identities and debate mainstream cultural norms. We are both experiencers and propagators, simultaneously active creators and meaningful manipulators. This has cultivated a vast and vibrant atmosphere and landscape, becoming the main driving force behind the creation of global culture and a stalwart backbone propelling cultural evolution. This repeatedly serves as a reminder that we ought to pay greater interest to expressions embedded within subcultures.

Specifically, rooted in the dynamism and ubiquity of subcultures (Amit & Wulff, 2022; De Kloet & Fung, 2016), our decision to commence with the spoonerism phenomenon expression stems from its historical prominence and persistent appeal (Bennett &

Kahn-Harris, 2020). Reverend William Archibald Spooner's linguistic gaffes have transcended time, evolving into a treasured form of linguistic amusement. This choice encapsulates the enduring legacy of language manipulation, resonating across generations. It is often a way of expressing by oneself by exchanging subjects, objects, or word collocations (Yule, 2022). For instance, "John Watson, you discover blind spots" would be creatively altered to "Blind spots, you discover John Watson," thereby introducing a dramatic twist.

Within spoonerism phenomenon expression, a fascination with the exchange of linguistic elements becomes evident, naturally, which closely parallels the concept of inversion rhetoric expression (Elton, 2016; Fitz & Chang, 2017). Inversion rhetoric expression frequently assumes the role of a textual manipulation within subcultures, with historical roots dating back to ancient poetry, and it also has persisted as a valuable literary device through to modern literature. For instance, the expression "Enfolding sunny spots of greenery" can be reordered to "Enfolding greenery spots of sunny." where the order of words or phrases in a sentence is reversed to create a rhetorical effect. This text reshuffling renders the overall semantics more striking and expressive, hence offering enhanced ornamental value. It is noteworthy that, the difference between inversion rhetoric expression and spoonerism phenomenon expression is that the former is more casual although both they involve the manipulation of word order.

Also, in the dynamic landscape of cultural exchange and the convergence of diverse languages, we observe that the boundaries between languages blur, giving rise to a unique linguistic crossroads. Within this context emerges the homophone meme expression (Schmidt & de Kloet, 2017; Wong et al., 2021). It involves the application of text fragments phonetically resembling Chinese text fragments, aiming to imbue the communication with a jovial yet subversive nuance. To illustrate, the Chinese phrase "贪生怕死 (cravenly cling to life instead of braving death)" can be artfully replaced with "贪生 pass (the pronunciation in Chinese means fear of death)." While this substitution preserves the phonetic semblance, it introduces a stronger negative and revolting undertone, augmenting the impact of the original expression.

Moreover, as the widespread use of emoji continues to flourish, a fascinating trend has emerged within subcultures—emoji encryption expression (Franco & Fugate, 2020), the practice of conveying information with an almost cryptic twist through the medium of emoji. Emoji encryption expression pertains to the playful substitution of main collocations within a text with emojis, thereby imparting an encryption-like effect. For instance, the phrase "beauty and the beast" might be whimsically encoded as 👩 and 🐗."

In parallel with the practice of encrypting collocations, there exists an equally captivating subcultural expression that operates in a rather deconstructed manner. Termed decomposed expression (Chen, 2021), this intriguing phenomenon centers on the art of breaking down language constructs into their elemental forms, the disassembly of a character/word into its sub-parts/subwords, which then serves as a breeding ground for the creation of fresh words or phrases possessing satirical and humorous connotations. For instance, the word "homesick" might be deconstructed into "home sick," another expression of negative sentiment.

At last, in today's fast-paced world, where information comes at us from all directions and our time is often fragmented, a subcultural expression has emerged that perfectly aligns with this era of mobile connectivity. Termed mobile data economizing expression (Zea & Heekyoung, 2019), it primarily centers on the art of concisely summarizing the essence of substantial content using a minimalistic approach, achieving the dual purpose of saving mobile data and capturing the audience's attention in a world characterized by information overload. An example might be the description "don't drink too much water at night" which encapsulates the lengthy discourse of science popularization personnel on this.

Considering the status that the model enjoys in informing human decision-making, it can be advantageous to augment its training with expressions derived from diverse subcultures, which potentially broadens its comprehension of nuanced human sentiments to enhance its performance.

## 3 | METHODOLOGY

The proposed SCDA consists of word collocation recognizer and subcultural expression generators, as shown in Figure 1. A word collocation recognizer is designed, termed BertRank, which can recognize Chinese and English word collocations. Viewed from the model's vantage point, it is beneficial to allow it to observe and assimilate knowledge from each sample through the lens of various perspectives, thereby diversifying its learning. SCDA now covers various typical expressions and meticulously engineers corresponding generators, namely, spoonerism phenomenon expression generator (SPEG), homophone meme expression generator (HMEG), emoji encryption expression generator (EEEG), inversion rhetoric expression generator (IREG), decomposed expression generator (DEG), and mobile data economizing expression generator (MDEEG). These generators serve the purpose of creating varied text expressions to enrich the data augmentation process for each individual text, thereby providing a more diverse dataset to combat insufficient volume.

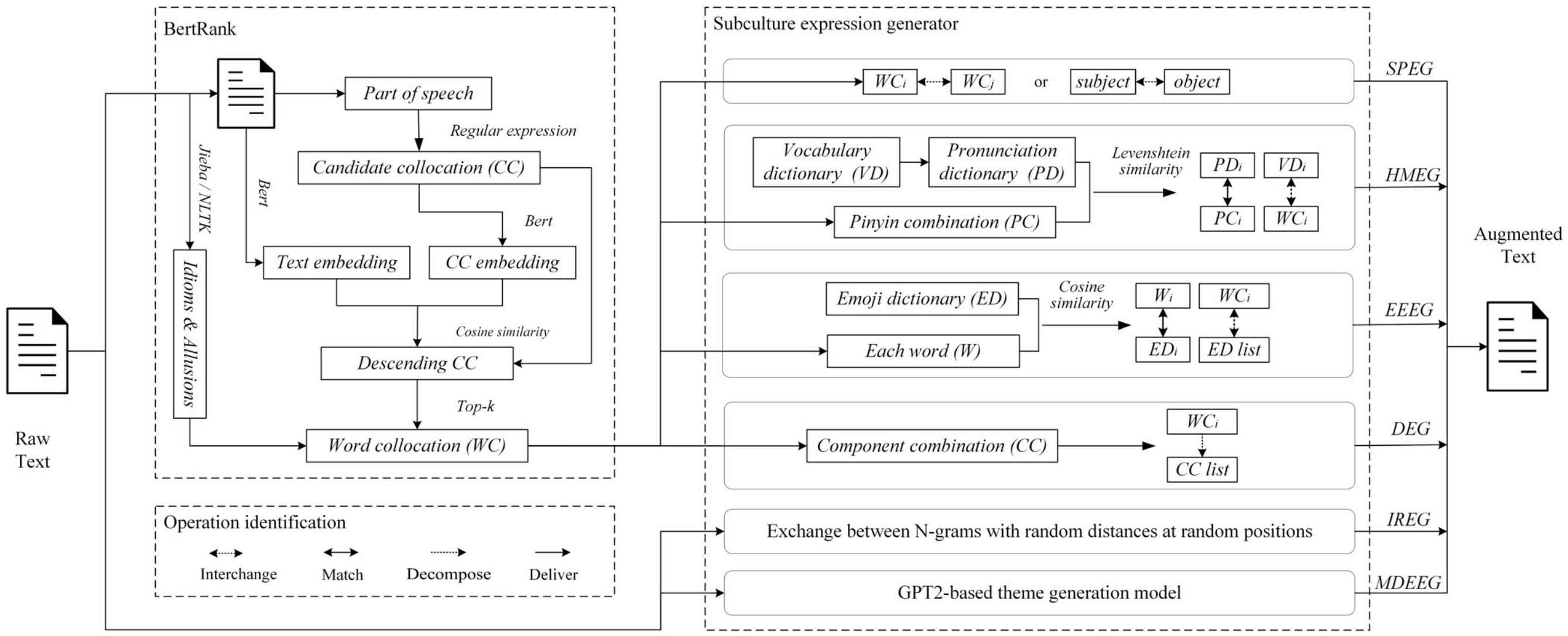

**FIGURE 1** Architecture of SCDA with diverse subcultural expression generators.

## 3.1 | Word collocation recognizer

Some of our generators require processing at the level of noun collocations. Since SIFRank (Song et al., 2023) is limited to English and relies on the less robust ELMO word representation, we optimize it into a Bert-based tool capable of recognizing both Chinese and English word collocations, termed BertRank.

Specifically, given a text $T$ formulated as multiple sentences with word sequence $\{w_1, w_2, ..., w_n\}$, BertRank is designed to discern primarily noun-based word collocations $WC$ within $T$. The first step involves utilizing an open-source NLP tool *jieba* (or *NLTK*) for part-of-speech (POS) tagging on $T$. The second step is to employ a rule template, written in regular expressions, to identify candidate word collocations $CC = \{cc_1, cc_2, ..., cc_m\}$, mainly noun combinations, and adjectives-noun pairs. The third step uses Bert to transform both the $CC$ and $T$ into their respective embeddings, $E_{cc}$ and $E_T$. The fourth step calculates the cosine distance between $E_{cc}$ and $E_T$, as the similarity, and top-k similarity-based $CC$ are selected to constitute the final set of word collocations $WC_1$.

In an effort to preserve word collocations comprehensively, BertRank employs *jieba* and so forth, as aid to extract the set $WC_2$, encompassing idioms and allusions (if any) from T. Hence, the resulting word collocations $WC$ represent a union of both $WC_1$ and $WC_2$.

## 3.2 | Spoonerism phenomenon expression generator

The spoonerism phenomenon expression often achieves humor by swapping the subject and object, or rearranging word collocations within a sentence. The SPEG is proposed, with the raw text $T$ as input and $T'$ as output. SPEG divides $T$ into multiple sentences $\{S_1, ..., S_q\}$ based on comma delimiters. For each sentence $S$ composed of words $\{w_1, ..., w_{s1}, w_{s2}, ..., w_{o1}, w_{o2}, ..., w_n\}$ of length $N$ or word collocations $\{wc_1, ..., wc_i, ..., wc_j, ..., wc_m\}$ of length $M$, SPEG employs a grammar dependency analysis tool to identify the subject and object and proceeds to interchange them if present, thereby generating $T' = \{w_1, ..., w_{o1}, w_{o2}, ..., w_{s1}, w_{s2}, ..., w_n\}$. Otherwise, BertRank is utilized to detect word collocations and perform a random exchange between them, generating $T' = \{wc_1, ..., wc_j, ..., wc_i, ..., wc_m\}$.

## 3.3 | Homophonic meme expression generator

The concept of a homophonic meme pertains to the substitution within Chinese/English text $T$ of word collocations $WC$ with English/Chinese combinations that exhibits a similar pronunciation, thereby forming an altered version $T'$. In light of this, the HMEG is proposed.

Provided with a Chinese text $T$ with multi-words $W = \{w_1, ..., w_i, ..., w_n\}$, HMEG initially utilizes BertRank to identify $WC = \{wc_1, ..., wc_i, ..., wc_m\}$. Subsequently, Python *xpinyin* library is employed to transcribe each $wc_i$ into Pinyin $P = \{p_1, p_2, ..., p_n\}$ for its character, and then combine $P$ based on its original arrange to form Pinyin combination $PC = \{p_1p_2, p_1p_2p_3, ..., p_kp_q, ...\} = \{pc_1, ..., pc_i, ..., pc_f\}$. Next, a vocabulary dictionary $VD = \{vd_1, ..., vd_i, ..., vd_r\}$ is prepared, where $VD$ considers common vocabulary with lengths ranging from 3 to 7 letters, ensuring syllable counts within the 2–3 range, as syllables empirically exceeding 4 tend to distort homophonic memes. This measure ensures readers recognize these words, such as "book" and so forth. By consulting $VD$, all word phonetic symbols are manually assigned corresponding pronunciations to form a pronunciation dictionary $PD = \{pd_1, ..., pd_i, ..., pd_g\}$. For example, the phonetic symbol "aɪ" corresponds to the Pinyin "ai," while "ð" resembles the pinyin "zhe," and stress markers such as "ˈ" are ignored. Finally, HMEG calculates the similarity between $pc_i$ and $pd_j$, and matches them accordingly. Given that the lengths of $pc_i$ and $pd_j$ often vary, the Levenshtein distance is employed as a similarity measure due to its suitability for strings of unequal lengths. The $vd_i$ with top-1 $pd_i$ versus $pc_j$ similarity is then selected to replace words $w_j^{pc}$ represented by $pc_j$, thereby forming the final $T' = \{w_1, ..., w_j^{pc}, ..., w_n\}$ in this way. Similarly, the substitution of English word collocations by Chinese follows the same logic.

## 3.4 | Emoji encryption expression generator

The notion of emoji encryption expression pertains to the representation of word collocations $WC$ in text $T$ by substituting them with emojis that bear similar literal meanings, thereby forming an altered text $T'$. With this in mind, the EEEG is proposed. EEEG commences by crawling emoji information from a website called *Emojiall*, constructing an emoji dictionary, $ED = \{ed_1, ..., ed_i, ..., ed_n\}$, which encapsulates the respective meanings of its elements, such as an emoji 🐗 representing "beast". Subsequently, EEEG employs BertRank to identify $WC = \{wc_1, ..., wc_j, ..., wc_m\}$, each containing a specific word count, $L_j$. For $wc_j$, EEEG matches the meaning of each word with an emoji. The employed similarity calculation is the Bert similarity measure (Tracz et al., 2020), since it is essential to amalgamate their semantics. The $wc_j$ is then replaced with $L_j$ $ed_i$ yielded by the top-1 similarity. This process continues until all $WC$ have been replaced by emojis, thereby formulating the final altered text $T'$.

## 3.5 | Inversion rhetoric expression generator

Inverted rhetoric expression often denotes the intent to emphasize and bolster semantics through the disruption of text structure. It presents in a multitude of forms, and its transformations are based on characters of varying lengths. In response, our proposed IREG utilizes a blend of random and Gaussian distributions. Specifically, the positions of the $n$-gram word segments are interchanged randomly with a random probability, and the distance of the exchange positions is derived from a Gaussian distribution. To illustrate, 76.4% of the adjacent word segments are swapped, 21.8% are separated by one character, 1.8% are separated by two characters, and so on. By employing this method, each text can generate an augmented text.

## 3.6 | Decomposed expression generator

Decomposed expression primarily refers to the substitution of word collocations $WC$ in text $T$ with their corresponding components, thereby forming an enhanced version, $T'$. In light of this, the DEG is proposed. BertRank initially identifies $WC = \{wc_1, ..., wc_i, ..., wc_n\}$. Then the Python *cnradical* library for Chinese and subword byte pair encoding for English are employed to break down each word of $wc_i$ into its components, and form a new sequence that replaces $wc_i$, thus generating $T'$. This process continues until all $WC$ have been transformed.

## 3.7 | Mobile data economizing expression generator

Mobile-data-economizing expression encompasses the creation of a concise text $T'$, which provides a high-level substitute of the original text $T$ and can naturally serve as enhanced data. This concept is abstracted as a theme induction in this paper. To achieve this, we employ a well-established GPT2 as our MDEEG, for creating a succinct theme for the $T$. It is trained on a corpus from *Weibo* and *Twitter*, capturing diverse themes including comments, news dissemination, and event documentation, with each user-generated content naturally aligned with a labeled sample, making it an ideal training resource.

Specifically, we initially obtain raw data utilizing a crawler and embark on a data cleaning process to eliminate symbols such as "HTML" and "# #." Once cleaned, the data is consolidated, filtering out duplicate entries, data entries with fewer than 100 words, and those with fewer than 2 title words. As a result, we procure train and test sets containing 24,000 and 3000 data entries, respectively.

For the convenience of training, we process the train set to separate each content and its theme using "SEP,"

terminated by "EOS." We leverage the Bert Tokenizer to generate token embeddings from the processed train set (Wang et al., 2023), simultaneously forming position embeddings based on each token's location. Moreover, to distinguish between the Content and Theme, SEP is embedded, serving as segment embeddings. The concatenation of these three is fed into a Transformer decoder with multiple parallel layers and 12 parallel heads in its attention mechanism. An autoregressive method is applied for sequence prediction, utilizing a cross-entropy loss function to minimize the loss value pertaining to the theme component. Once trained over 10 epochs, MDEEG can generate a succinct theme of maximum 32 characters, serving as a mechanism of data enhancement.

### 3.8 | Section summary

In a concerted effort to foster data augmentation for sentiment analysis, we propose the SCDA approach, devising distinct subculture-based expression generators. Notably, each entry is enriched through the generation of distinct representation samples, imbuing the data with a tapestry of diverse forms. This multi-faceted way unlocks an unparalleled opportunity for observation and learning, enabling it to process each sample from additional perspectives. By immersing itself in this subculture landscape, the model is expected to develop an enhanced capability. In the forthcoming section, we embark on an extensive experimental evaluation to ascertain the effectiveness of SCDA. Our objective is to investigate the potential of SCDA in augmenting the performance of sentiment analysis models.

## 4 | EXPERIMENT

### 4.1 | Dataset

To conduct a comprehensive evaluation of our SCDA, we employ three publicly available sentiment analysis datasets: Chnsenticorp (Chen et al., 2015), Semeval (Rosenthal et al., 2019) and ACSA (Bu et al., 2021), https://github.com/wzh-ins/sentiment-analysis. The Chnsenticorp dataset comprises hotel reviews expressing positive and negative sentiments, while the Semeval dataset focuses on Twitter, involving positive and negative sentiments from social media.

The ACSA dataset offers a more fine-grained evaluation. It focuses on restaurant reviews and provides sentiments (positive, neutral, and negative) across 18 distinct aspects, such as location, service, price, and ambience, see Table 1. This multi-aspect dataset enables us to comprehensively assess the performance of sentiment analysis models.

We randomly select 2000 samples from both datasets for our experiments, thus allowing us to explore the performance of SCDA in situations where data volume is insufficient, highlighting the potential of our approach in overcoming data scarcity challenges.

### 4.2 | Experiment setting

Four representative models are selected for evaluation: TextRNN, Transformer (Cunha et al., 2023), Bert with size of base (Pérez Pozo et al., 2022; Wang et al., 2022; Wang et al., 2024) and LLAMA2 with size of 7B (Touvron et al., 2023). These models, widely acclaimed and adopted, collectively embody distinct stages in the progression of deep learning, presenting a rich diversity. Particularly noteworthy is LLAMA2, which can be regarded as the latest generation of large-scale model. It has undergone instruction fine-tuning on the basis of LoRA (Hu et al., 2021), and the constructed prompt's instruction is "Please determine whether the following content expresses a positive sentiment, and output 0 or 1 - ->," with specific content included as input, corresponding labels as output. In contrast, the comparative methods encompass the commonly used techniques such as DICT

TABLE 1 Information on ACSA samples.

| Aspects | Positive | Neutral | Negative |
|---|---|---|---|
| Food#Taste | 1096 | 717 | 97 |
| Food#Appearance | 382 | 94 | 62 |
| Food#Portion | 524 | 167 | 182 |
| Food#Recommend | 31 | 34 | 92 |
| Price#Level | 306 | 454 | 235 |
| Price#Cost effective | 365 | 41 | 55 |
| Price#Discount | 311 | 302 | 41 |
| Location#Downtown | 331 | 6 | 7 |
| Location#Transportation | 353 | 12 | 18 |
| Location#Easy to find | 272 | 36 | 58 |
| Service#Queue | 93 | 127 | 97 |
| Service#Hospitality | 673 | 290 | 165 |
| Service#Parking | 71 | 31 | 29 |
| Service#Timely | 158 | 52 | 106 |
| Ambience#Decoration | 653 | 198 | 40 |
| Ambience#Noise | 388 | 114 | 81 |
| Ambience#Space | 399 | 204 | 120 |
| Ambience#Sanitary | 465 | 117 | 99 |

**TABLE 2** Evaluation results.

| Models | ACSA | | Chnsenticorp | | Semeval | |
|---|---|---|---|---|---|---|
| | Val | Test | Val | Test | Val | Test |
| TextRNN | 63.11 | 63.34 | 80.36 | 80.07 | 66.25 | 66.33 |
| TextRNN + SCDA | **69.31** | **69.98** | **81.65** | **81.38** | **68.60** | **69.04** |
| Transformer | 68.31 | 68.90 | 81.63 | 80.60 | 70.43 | 70.11 |
| Transformer + SCDA | **72.96** | **72.91** | **82.28** | **82.54** | **72.09** | **71.82** |
| Bert | 76.45 | 77.35 | 84.07 | 84.63 | 76.01 | 74.55 |
| Bert + SCDA | **78.88** | **78.74** | **87.14** | **86.77** | **78.82** | **77.60** |
| LLAMA2 | 94.59 | 95.47 | 94.24 | 95.15 | 96.13 | 96.28 |
| LLAMA2 + SCDA | **96.20** | **97.53** | **96.91** | **97.42** | **97.94** | **98.47** |

*Note:* Bold values signifies as the best performance in each set of experiments.

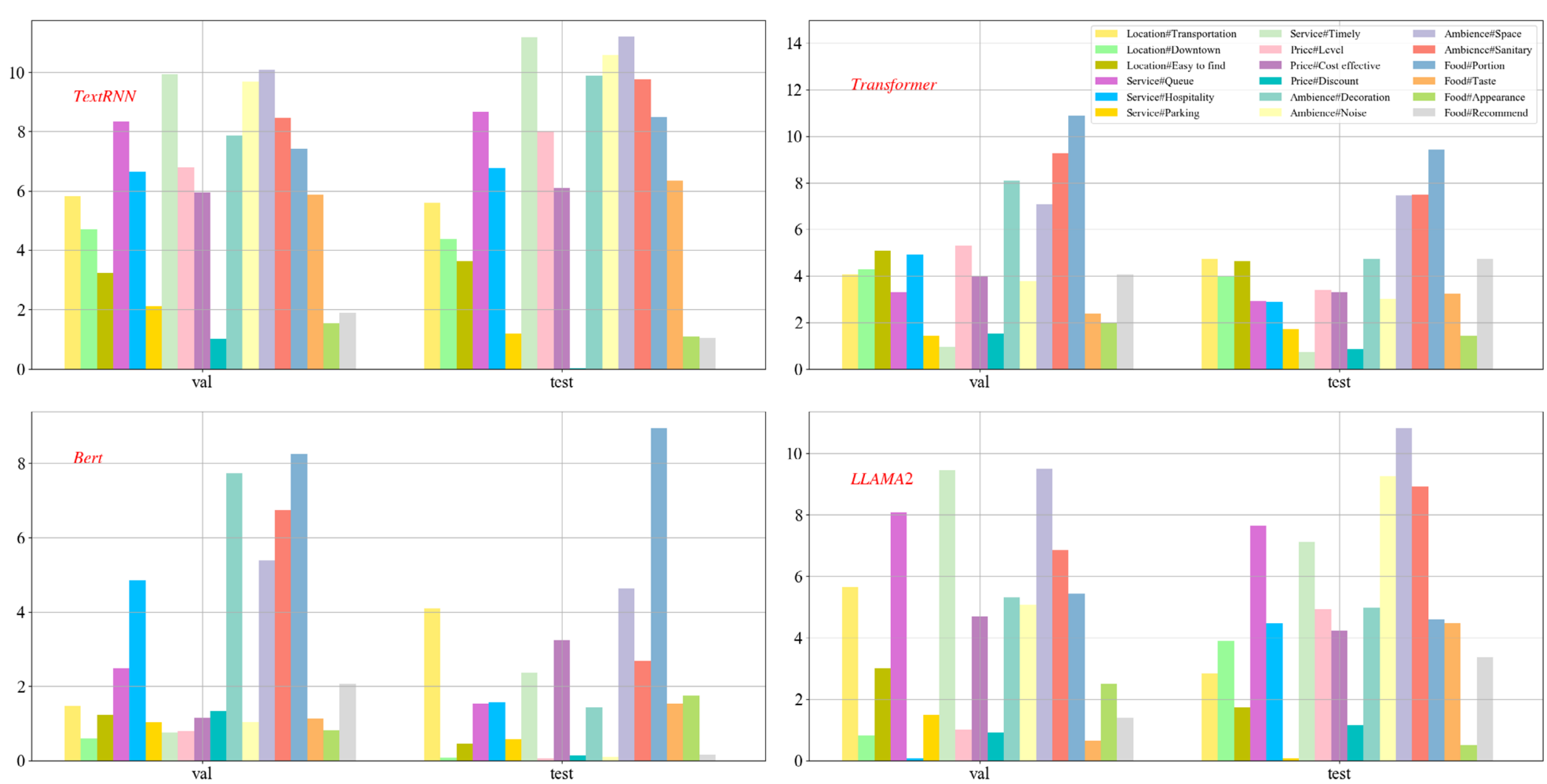

**FIGURE 2** Performance gains on various aspects of ACSA.

(Zhang et al., 2015), EDA (Wei & Zou, 2019), and PLSDA (Xiang et al., 2021). The metrics is the accuracy (%).

## 5 | RESULTS

### 5.1 | Evaluation experimental results

The evaluation results are summarized in Table 2. The findings from the validation set ("val") and test set ("test") provide evidence for the advantages conferred by incorporating SCDA into the four models.

TextRNN demonstrates a substantial augmentation in its prowess upon ACSA, with an increase surpassing 6%. Moreover, an approximate elevation of 1%–3% has been discerned in Chnsenticorp and Semeval. Noteworthily, Transformer shows an improvement of over 4% within ACSA, with noticeable enhancements radiating across the remaining two datasets. Furthermore, Bert evinces a performance gains of approximately 2% to 3% on all three datasets, notably evident in Chnsenticorp's validation set and Semeval's test set. Simultaneously, an enhancement of 1% to 2% has been attained upon LLAMA2. These results accentuate the effectiveness of SCDA.

Figure 2 highlights the remarkable performance gains across all aspects within ACSA, attributed to the formidable influence of SCDA, wherein the vertical axis quantifies the gains. Specifically, Ambience#Space and

TABLE 3 Comparison results of SCDA with DICT, EDA, and PLSDA.

| Models | ACSA | | Chnsenticorp | | Semeval | |
|---|---|---|---|---|---|---|
| | Val | Test | Val | Test | Val | Test |
| TextRNN + DICT | 64.25 | 64.17 | 80.95 | 80.77 | 66.51 | 67.03 |
| TextRNN + EDA | 67.44 | 66.86 | 81.02 | 80.94 | 67.32 | 67.86 |
| TextRNN + PLSDA | 68.58 | 68.93 | 81.17 | 80.85 | 67.74 | 68.27 |
| TextRNN + SCDA | **69.31** | **69.98** | **81.65** | **81.38** | **68.60** | **69.04** |
| Transformer + DICT | 70.33 | 70.62 | 81.79 | 81.23 | 70.88 | 70.53 |
| Transformer + EDA | 70.52 | 71.03 | 81.81 | 81.88 | 71.30 | 70.84 |
| Transformer + PLSDA | 70.48 | 71.15 | 82.05 | 82.02 | 71.75 | 71.01 |
| Transformer + SCDA | **72.96** | **72.81** | **82.28** | **82.54** | **72.09** | **71.82** |
| Bert + DICT | 76.93 | 77.62 | 84.86 | 85.01 | 76.34 | 74.91 |
| Bert + EDA | 77.12 | 77.90 | 85.43 | 85.52 | 76.93 | 75.85 |
| Bert + PLSDA | 77.96 | 78.25 | 86.28 | 86.32 | 77.46 | 76.67 |
| Bert + SCDA | **78.88** | **78.74** | **87.14** | **86.77** | **78.82** | **77.60** |
| LLAMA2 + DICT | 94.66 | 95.58 | 94.57 | 95.33 | 92.16 | 92.39 |
| LLAMA2 + EDA | 94.87 | 95.69 | 94.92 | 95.85 | 92.60 | 92.85 |
| LLAMA2 + PLSDA | 95.03 | 95.95 | 94.84 | 96.26 | 92.78 | 93.41 |
| LLAMA2 + SCDA | **96.20** | **97.53** | **96.91** | **97.42** | **97.94** | **98.47** |

*Note:* Bold values signifies as the best performance in each set of experiments.

Ambience#Sanitary have greatly benefited, primarily with an augmentation exceeding 7%. Ambience#Noise and Ambience#Decoration also exhibit noteworthy increases. Interestingly, the advancements exhibited by the quartet of models in Food#Portion remain steadfastly elevated. Yet, certain aspects such as Service#Parking, Price#Discount and Food#Recommend exhibit more modest gains, indicating that enriching raw expressions associated with these aspects through SCDA might pose challenges. This could stem from the limitations in BertRank's ability to capture relevant information and the constrained capabilities of generators, possibly hindering the model's exploration of additional perspectives.

## 5.2 | Comparison experimental results

Table 3 presents the comparison experimental results (total aspects), which demonstrate the competitiveness of SCDA by yielding higher gains compared to DICT, EDA and PLSDA.

SCDA does bestow upon models a greater wealth of stimulation, allowing them to navigate through diverse sentiments. Within the expansive subcultural expressions, models encounter various sentiment states, which may embolden models to transcend conventional boundaries, thereby fostering innovative surprises. Conversely, DICT, EDA, and PLSDA appear to tread a path of relative conservatism. For example, in the phrase "A stirring and funny re-imagining of beauty and the beast," DICT might substitute "beauty" with words like "lovely" or "eyeful." EDA could insert, exchange, or delete random words, resulting in "A stirring blessing of beauty re-imagining and the beast." PLSDA primarily pursues grammatical consistency, yielding instances like "A stirring and amusing re-imagining of beauty and the animal."

While these methods often rely on lexical substitution strategies, they might constrain the model within a relatively unchanging and creativity-starved training environment, akin to the mechanistic and stereotypic learning. While regimented learning undeniably yields fruit in specific contexts, it bears the potential to asphyxiate the model's potential for expansive applicability and imaginative evolution.

In contrast, SCDA involves currents of expression vitality in contemporary society, offering greater openness and creative liberty. For instance, in the phrase "A stirring and funny re-imagining of 👩 and 🐗," the emojis convey cute and beautiful female characters and symbolic representations related to strength, and ferocity respectively.

Undoubtedly, SCDA embodies a daring spirit, and the generated expressions carry the risk of both breakthroughs and counterproductive outcomes, especially if a model's understanding of sentiments within subcultures is limited. This phenomenon becomes evident on

TABLE 4 Results of ablation experiments.

| | ACSA | | Chnsenticorp | | Semeval | |
|---|---|---|---|---|---|---|
| | **Val** | **Test** | **Val** | **Test** | **Val** | **Test** |
| TextRNN + SCDA | | | | | | |
| All | 69.31 | 69.98 | 81.65 | 81.38 | 68.60 | 69.04 |
| Without SPEG | 68.97 (−0.34) | 69.07 (−0.91) | 81.22 (−0.43) | 80.70 (−0.68) | 67.82 (−0.78) | 67.84 (−1.20) |
| Without HMEG | 69.08 (−0.23) | 69.33 (−0.65) | 80.61 (−1.04) | 80.29 (−1.09) | 67.74 (−0.86) | 68.13 (−0.91) |
| Without EEEG | 69.14 (−0.17) | 69.21 (−0.77) | 80.53 (−1.12) | 80.16 (−1.22) | 67.95 (−0.65) | 68.82 (−0.22) |
| Without IREG | 68.92 (−0.39) | 68.95 (−1.03) | 80.87 (−0.78) | 80.80 (−0.58) | 67.62 (−0.98) | 67.99 (−1.05) |
| Without DEG | 69.05 (−0.26) | 69.46 (−0.52) | 80.85 (−0.80) | 81.21 (−0.17) | 68.52 (−0.08) | 68.18 (−0.86) |
| Without MDEEG | **69.77** (+0.46) | 69.94 (−0.04) | 81.14 (−0.51) | 80.91 (−0.47) | 68.09 (−0.51) | 68.54 (−0.50) |
| Transformer + SCDA | | | | | | |
| All | 72.96 | 72.81 | 82.28 | 82.54 | 72.09 | 71.82 |
| Without SPEG | 72.33 (−0.63) | 72.42 (−0.39) | 82.25 (−0.03) | 81.76 (−0.78) | 71.83 (−0.26) | 71.73 (−0.09) |
| Without HMEG | 71.99 (−0.97) | 71.97 (−0.84) | 81.25 (−1.03) | 81.41 (−1.13) | 70.97 (−1.12) | 70.62 (−1.20) |
| Without EEEG | 72.52 (−0.44) | 72.36 (−0.45) | 81.79 (−0.49) | 81.35 (−1.19) | 71.94 (−0.15) | 71.20 (−0.62) |
| Without IREG | 72.12 (−0.84) | 72.14 (−0.65) | 81.34 (−0.94) | 81.67 (−0.87) | 71.23 (−0.86) | 70.76 (−1.06) |
| Without DEG | 72.45 (−0.51) | 72.33 (−0.48) | 81.78 (−0.50) | 82.48 (−0.06) | 71.84 (−0.25) | 71.67 (−0.15) |
| Without MDEEG | **73.23** (+0.27) | **73.00** (+0.19) | 81.86 (−0.42) | 82.24 (−0.30) | 72.06 (−0.03) | 71.63 (−0.19) |
| Bert + SCDA | | | | | | |
| All | 78.88 | 78.74 | 87.14 | 86.77 | 78.82 | 77.60 |
| Without SPEG | 78.74 (−0.14) | 78.38 (−0.36) | 86.38 (−0.76) | 86.18 (−0.59) | 78.50 (−0.32) | 77.27 (−0.33) |
| Without HMEG | 78.21 (−0.67) | 77.92 (−0.82) | 85.95 (−1.19) | 85.64 (−1.13) | 78.00 (−0.82) | 76.61 (−0.99) |
| Without EEEG | 78.41 (−0.47) | 77.84 (−0.90) | 86.72 (−0.42) | 86.40 (−0.37) | 78.62 (−0.20) | 77.39 (−0.21) |
| Without IREG | 78.27 (−0.61) | 77.58 (−1.16) | 86.04 (−1.10) | 86.24 (−0.53) | 77.69 (−1.13) | 76.43 (−1.17) |
| Without DEG | 78.79 (−0.09) | 78.18 (−0.56) | 86.83 (−0.31) | 86.53 (−0.24) | 78.39 (−0.43) | 76.87 (−0.73) |
| Without MDEEG | **78.99** (+0.11) | 78.61 (−0.13) | 86.96 (−0.18) | 86.51 (−0.26) | 78.67 (−0.15) | 77.04 (−0.56) |
| LLAMA2 + SCDA | | | | | | |
| All | 96.20 | 97.53 | 96.91 | 97.42 | 97.94 | 98.47 |
| Without SPEG | 95.91 (−0.29) | 97.45 (−0.08) | 96.14 (−0.77) | 96.55 (−0.87) | 97.29 (−0.65) | 97.65 (−0.82) |
| Without HMEG | 96.07 (−0.13) | 96.73 (−0.80) | 96.62 (−0.29) | 97.13 (−0.29) | 97.15 (−0.79) | 97.86 (−0.61) |
| Without EEEG | 95.35 (−0.85) | 96.56 (−0.97) | 95.74 (−1.17) | 96.38 (−1.04) | 96.78 (−1.16) | 97.60 (−0.87) |
| Without IREG | 96.03 (−0.17) | 96.77 (−0.76) | 96.80 (−0.11) | 96.71 (−0.71) | 96.90 (−1.04) | 97.55 (−0.92) |
| Without DEG | 95.80 (−0.40) | 97.46 (−0.07) | 96.27 (−0.64) | 96.31 (−1.11) | 97.77 (−0.17) | 97.77 (−0.70) |
| Without MDEEG | 96.02 (−0.18) | 97.33 (−0.20) | 95.99 (−0.92) | 96.46 (−0.96) | 97.47 (−0.47) | 97.78 (−0.69) |

*Note:* Bold values signifies as the best performance in each set of experiments.

Chnsenticorp, where models like TextRNN do not significantly outperform previous methods, while large language model like LLAMA2 demonstrate more substantial performance enhancements, owing to its greater comprehension and learning capacity.

Viewed from an alternative perspective, a closer alignment between the constructed text and the original could, to a certain extent, indicate that the model regards such augmentations as burdensome, potentially resulting in uninspired outputs. This could be observed in cases involving LLAMA2, a model well-versed in natural language operations like word substitution and insertion as seen in DICT and EDA, while also placing importance on the grammatical norms emphasized by PLSDA. Consequently, data produced through these methods might appear somewhat repetitive, serving to reinforce impressions but lacking the potential to significantly enhance the model's performance. From this vantage point, SCDA

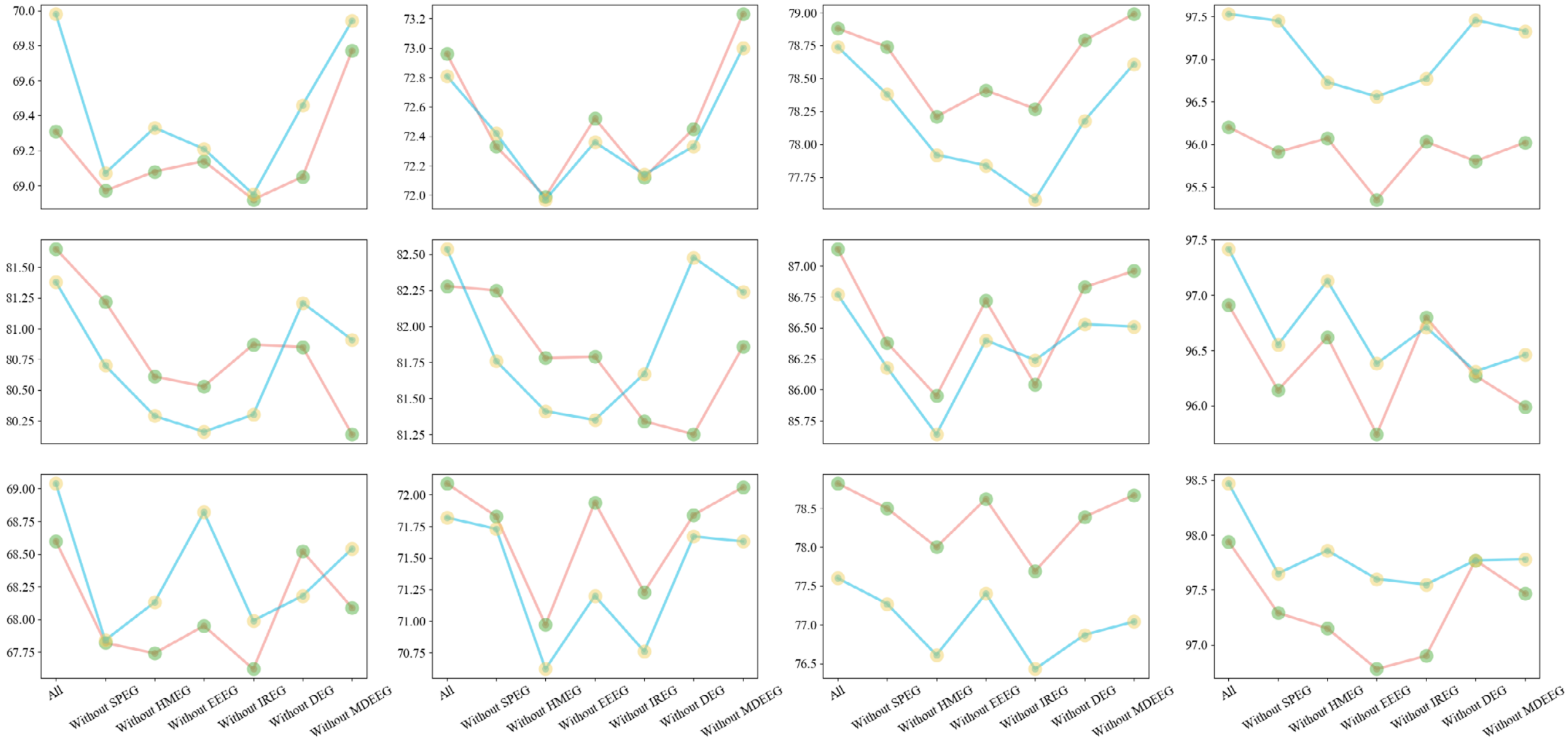

**FIGURE 3** The fluctuation amplitude of performance. Red lines: validation set, blue ones for test set. Row 1–3: ACSA, Chnsenticorp and Semeval. Column 1–4: TextRNN, Transformer, Bert and LLAMA2.

enjoys a distinct advantage, as it offers supplementary inspiration for the model.

## 5.3 | Ablation experimental results

To assess the influence of each generator on SCDA, we conducted ablation experiments wherein we exclude each generator (SPEG, HMEG, EEEG, IREG, DEG, and MDEEG) individually. The results (total aspects) are summarized in Table 4, accompanied by the corresponding performance fluctuations denoted in parentheses. Figure 3 illustrates the variations observed during the ablation experiments.

On ACSA, the performance of TextRNN's significantly drops with the absence of SPEG and IREG, indicating their substantial impact on the model. For Transformer and Bert, HMEG and IREG stand out as more influential stimuli. Remarkably, IREG exerts a stronger influence on both models, potentially attributed to the historical roots of inversion rhetoric expressions in the language and culture. These expressions evolved from maintaining rhythmic harmony in ancient poetry and miscellaneous music, to creating diverse aesthetic in modern prose, and eventually finding a place in contemporary subcultures to emphasize sentiments in pursuit of "novelty". This natural coherence with human cognition allows for effortlessly comprehension and integration into communication. Chnsenticorp, to some extent, maintains stimulation from IREG while highlighting the impressive capabilities of HMEG and EEEG. This emphasis aligns with Chnsenticorp's focus on the single-dimensional polarity of sentiment, using homophones and emojis to accentuate expressions. Semeval's results echo the above findings and reflect the surprising influence of IREG, especially for Transformer and Bert, while acknowledging the notable performance of HMEG, and generally commendable performance of EEEG.

Interestingly, LLAMA2 exhibits a distinct preference for EEEG over the counterparts. The absence of EEEG notably affects LLAMA2's performance, particularly in the validation sets across all three datasets. This could be due to LLAMA2's robust natural language understanding, making it less sensitive to textual manipulation and more drawn to the reservoir of sentiments encapsulated therein. This phenomenon, in turn, underscores the progressive nature of SCDA compared with previous methods.

Interestingly, TextRNN, Transformer and Bert all exhibit performance gains on ACSA when MDEEG is excluded. This could be attributed to a potential contradiction with the original text, as the mobile data economizing expressions pertains to themes generated by GPT-2. These themes are relatively short in length, which may lead to biased interpretations or inevitable omissions of information. Our investigation into the similarity between subcultural expressions and the original text reveals that the mobile data economizing expressions exhibit a relatively lower mean similarity of 0.74–0.80 (with a standard deviation of 0.01–0.04) compared to other expressions. This, to some extent, explains why MDEEG has hindered the models, as shown in Figure 4.

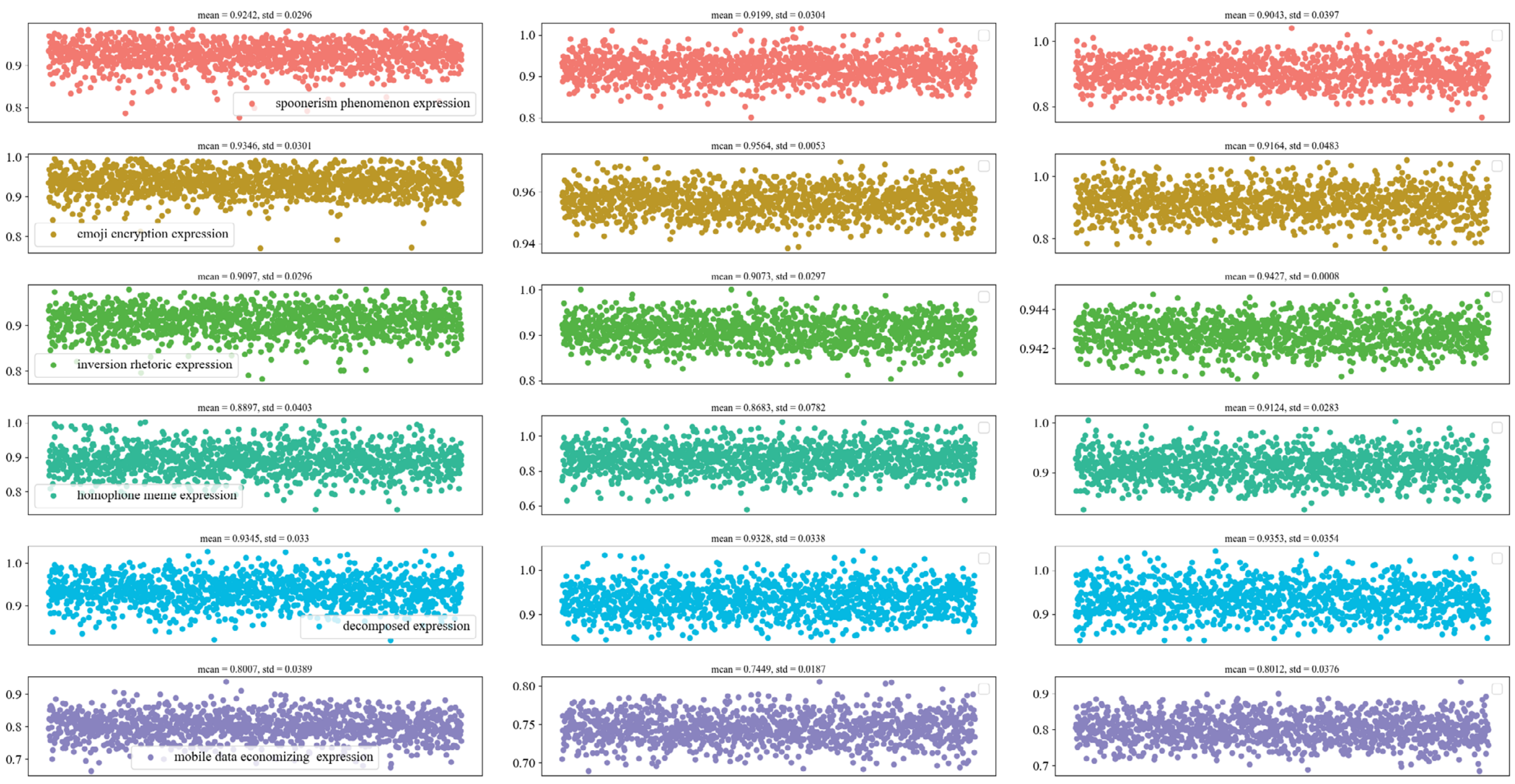

**FIGURE 4** Similarity between subcultural expression and original text on ACSA, Chnsenticorp and Semeval from left to right.

**TABLE 5** Results on mobile data economizing expressions (MDEE) themselves versus raw text.

| Models/datasets | ACSA | | Chnsenticorp | | Semeval | |
|---|---|---|---|---|---|---|
| | **Val** | **Test** | **Val** | **Test** | **Val** | **Test** |
| TextRNN → raw text | 63.11 | 63.34 | 80.36 | 80.07 | 66.25 | 66.33 |
| TextRNN → MDEE | 51.24 (−11.87) | 53.09 (−10.25) | 76.45 (−3.91) | 75.33 (−4.74) | 62.05 (−4.20) | 62.14 (−4.19) |
| Transformer → raw text | 68.31 | 68.90 | 81.63 | 80.60 | 70.43 | 70.11 |
| Transformer → MDEE | 57.97 (−10.34) | 57.65 (−11.25) | 76.71 (−4.92) | 76.52 (−4.08) | 65.89 (−4.54) | 66.97 (−3.14) |
| Bert → raw text | 76.45 | 77.35 | 84.07 | 84.63 | 76.01 | 74.55 |
| Bert → MDEE | 68.62 (−7.83) | 69.16 (−8.19) | 79.53 (−4.54) | 79.82 (−4.81) | 72.48 (−3.53) | 69.62 (−4.93) |
| LLAMA2 → raw text | 94.59 | 95.47 | 94.24 | 95.15 | 96.13 | 96.28 |
| LLAMA2 → MDEE | 90.47 (−4.12) | 90.83 (−4.64) | 92.04 (−2.20) | 91.39 (−3.79) | 93.72 (−2.41) | 93.06 (−3.22) |

Moreover, considering the unexpected trend where MDEEG enhances the model's performance on Chnsenticorp and Semeval, we conjecture that ACSA, given its wide granularity and multi-faceted polarity, is sensitive to the potential information omissions or biases in the mobile data economizing expression, thus limiting the model's ability for nuanced sentiment analysis. Conversely, Chnsenticorp and Semeval, with their singular focus on binary positive and negative polarities, benefit from MDEEG's thematic induction behavior in the training text, offering greater clarity and a more conducive environment for model training. To verify this idea, we conduct experiments on the enhanced data of MDEE, as shown in Table 5. We observe that the impact of MDEE on model learning in Chnsenticorp and Semeval closely aligns with the original text, with performance differences fluctuating at approximately 4%. In contrast, ACSA exhibits more significant performance differences, notably around 11% on TextRNN and Transformer. These findings provide empirical evidence supporting our notion.

In conclusion, SCDA, has proven to be effective through the innovative integration of subcultural expressions, significantly enhancing sentiment analysis.

## 6 | DISCUSSION

The successful implementation of SCDA provide a valuable contribution to data augmentation and highlight the importance of exploring diverse linguistic and cultural phenomena to enhance the sentiment analysis models. The findings also suggest the value of incorporating linguistic variations into data augmentation techniques to better capture the nuances and intricacies of sentiment expression in different contexts. This study could have been enriched from several angles.

First, while acknowledging the constantly evolving nature of subcultures, it is challenging to encompass all emerging forms of expressions. Nonetheless, including various enduring aspects is a significant step forward, providing a foundation for studying subcultural expressions and enhancing sentiment analysis models. This approach provides a multifaceted view of subculture and sentiment, recognizing ongoing exploration and documentation of emerging expressions within this dynamic landscape.

Future research could expand the coverage by incorporating more diverse and specific subcultures, considering regional and cultural variations, and exploring emerging trends within subcultures. By continuously exploring and updating the generators, researchers can stay up-to-date with the evolving subculture landscape and provide more comprehensive data augmentation techniques. Additionally, it would be beneficial to validate the effectiveness of the proposed generators in sentiment analysis tasks through experiments and evaluations using various sentiment analysis models and datasets. This would help establish the practical applicability and generalizability of the generators in real-world sentiment analysis scenarios.

Second, through extensive investigations, we have an interesting observation regarding a reversible relationship between the spoonerism or inversion rhetoric expression, labeled as $T'$, and the original text $T$. This reversible relationship is observed through a vector representation, where a linear transformation $\psi$ exists between $T$ and $T'$. This observation brings to light a matrix equation defining the maximum non-repeating elements in $T$ and $T'$. To illustrate, we use the following text as an example.

$T =$ (John Watson, you discover blind spots).
$T' =$ (Blind spots, you discover John Watson).

We establishing a base vector representation for both. Obviously, both vectors have six elements. Let the maximum non-repeating element in $T$ and $T'$ be the base vector, that is, $e_1 =$ John; $e_2 =$ Watson; $e_3 =$ you; $e_4 =$ discover; $e_5 =$ blind; $e_6 =$ spots. There is a matrix equation. In this regard, there is a linear transformation $\psi$ that satisfies $T = \psi T'$.

$$T = \begin{bmatrix} 1 & & & & & \\ & 1 & & & & \\ & & 1 & & & \\ & & & 1 & & \\ & & & & 1 & \\ & & & & & 1 \end{bmatrix} \begin{bmatrix} e_1 \\ e_2 \\ e_3 \\ e_4 \\ e_5 \\ e_6 \end{bmatrix} = Ie,$$

$$T' = \begin{bmatrix} & & & & 1 & \\ & & & & & 1 \\ & & 1 & & & \\ & & & 1 & & \\ 1 & & & & & \\ & 1 & & & & \end{bmatrix} \begin{bmatrix} e_1 \\ e_2 \\ e_3 \\ e_4 \\ e_5 \\ e_6 \end{bmatrix} = Ae.$$

We perceive that the matrix is invertible, $A = A^{-1}$, indicating that $T = \psi T'$ and $T' = \psi T$ hold true. This sheds light on the inherent structure and logic behind these linguistic phenomena, paving the way for further exploration and improvement in this area. Future research can delve deeper into understanding the underlying mechanisms and implications of this linear relationship. By studying the matrix $A$, researchers can gain insights into the fundamental principles governing the generation and interpretation of these expressions. Exploring the invertibility of the matrix $A$ and its implications for language processing and comprehension can provide valuable insights into human communication and expression.

The notion that people express themselves based on the restorability of logic is an intriguing aspect to consider. This observation suggests that individuals, when using spoonerism or inversion rhetoric expressions, implicitly rely on the listener's ability to decode and reconstruct the original logical structure. Investigating the cognitive processes involved in decoding these expressions can lead to a better understanding of human language comprehension and the role of context in communication.

Future studies can explore several aspects. Studying the matrix $A$ in different linguistic contexts and languages could provide a more comprehensive understanding of the reversible nature of these expressions. Investigating the cognitive mechanisms and neural processes underlying the interpretation of reversible expressions could provide insights into the neural basis of language processing and comprehension. Furthermore, examining the potential applications of this discovery in NLP and sentiment analysis could contribute to the development of more sophisticated language models and sentiment analysis algorithms. Incorporating the insights from the reversible expressions could enhance the performance of sentiment analysis systems, improving their ability to interpret and understand subculture-based linguistic variations.

Finally, it is crucial to consider the ethical implications of using subcultural expressions in sentiment analysis. Subcultures often arise from specific communities or groups, and their expressions may carry sensitive or exclusive meanings. Potential biases or misinterpretations should be considered. Future research could incorporate ethical considerations, diversity, and inclusivity into the research process to ensure responsible representation of different subcultures.

## 7 | CONCLUSION

This study focuses on sentiment analysis within the emerging subcultures prevalent in today's online landscape, and explores whether the integration of subcultural expressions for data augmentation can enhance sentiment analysis models, particularly in scenarios with limited training data. In this regard, we devise distinct expression generators corresponding to specific subcultural expression categories, allowing for the generation of expanded text for training. This study employs Chnsenticorp, Semeval, and ACSA as datasets, and conducts experiments using the TextRNN, Transformer, Bert, and LLAMA2. The results provide comprehensive evidence of the effectiveness of the proposed approach and offer insightful findings from various perspectives. By augmenting the training data with subcultural expressions, sentiment analysis models could potentially develop a better understanding of the subtleties and intricacies within different subcultures, resulting in improved performance in sentiment analysis tasks. This study also uncovers a fascinating connection between subcultural expression and human speech cognition, paving the way for future exploration into the intricate interplay between information, emotions, and culture in future endeavors.

## ACKNOWLEDGMENTS

This work is supported by the Outstanding Innovative Talents Cultivation Funded Programs 2023 of Renmin Univertity of China.

## CONFLICT OF INTEREST STATEMENT

The authors declare no conflicts of interest.

## ORCID

*Zhenhua Wang* https://orcid.org/0000-0002-0369-2765

## REFERENCES

Abonizio, H. Q., Paraiso, E. C., & Barbon, S. (2021). Toward text data augmentation for sentiment analysis. *IEEE Transactions on Artificial Intelligence*, *3*(5), 657–668. https://doi.org/10.1109/tai.2021.3114390

Alqudah, R., Al-Mousa, A. A., Hashyeh, Y. A., & Alzaibaq, O. Z. (2023). A systemic comparison between using augmented data and synthetic data as means of enhancing wafermap defect classification. *Computers in Industry*, *145*, 103809. https://doi.org/10.1016/j.compind.2022.103809

Amit, V., & Wulff, H. (Eds.). (2022). *Youth cultures: A cross-cultural perspective*. Taylor & Francis. https://doi.org/10.4324/9781003333487

Bennett, A., & Kahn-Harris, K. (Eds.). (2020). *After subculture: Critical studies in contemporary youth culture*. Bloomsbury Publishing.

Bi, Y. (2022). Sentiment classification in social media data by combining triplet belief functions. *Journal of the Association for Information Science and Technology*, *73*(7), 968–991. https://doi.org/10.1002/asi.24605

Bu, J., Ren, L., Zheng, S., Yang, Y., Wang, J., Zhang, F., & Wu, W. (2021). ASAP: A Chinese review dataset towards aspect category sentiment analysis and rating prediction. *ACL*, 2069–2079. https://doi.org/10.18653/v1/2021.naacl-main.167

Bueno, I., Carrasco, R. A., Ureña, R., & Herrera-Viedma, E. (2022). A business context aware decision-making approach for selecting the most appropriate sentiment analysis technique in e-marketing situations. *Information Sciences*, *589*, 300–320. https://doi.org/10.1016/j.ins.2021.12.080

Chen, Z., Huang, Y., Tian, J., Liu, X., Fu, K., & Huang, T. (2015). Joint model for subsentence-level sentiment analysis with Markov logic. *Journal of the Association for Information Science and Technology*, *66*(9), 1913–1922. https://doi.org/10.1002/asi.23301

Chen, Z. T. (2021). Poetic prosumption of animation, comic, game and novel in a post-socialist China: A case of a popular video-sharing social media Bilibili as heterotopia. *Journal of Consumer Culture*, *21*(2), 257–277. https://doi.org/10.1177/1469540518787574

Chung, W., & Zeng, D. (2016). Social-media-based public policy informatics: Sentiment and network analyses of US immigration and border security. *Journal of the Association for Information Science and Technology*, *67*(7), 1588–1606. https://doi.org/10.1002/asi.23449

Cruz, N. P., Taboada, M., & Mitkov, R. (2016). A machine-learning approach to negation and speculation detection for sentiment analysis. *Journal of the Association for Information Science and Technology*, *67*(9), 2118–2136. https://doi.org/10.1002/asi.23533

Cunha, W., Viegas, F., França, C., Rosa, T., Rocha, L., & Gonçalves, M. A. (2023). A comparative survey of instance selection methods applied to nonneural and transformer-based text classification. *ACM Computing Surveys*, *55*, 1–52. https://doi.org/10.1145/3582000

De Kloet, J., & Fung, A. Y. (2016). *Youth cultures in China*. John Wiley & Sons.

Elton, W. R. (2016). *Shakespeare's Troilus and Cressida and the Inns of Court revels*. Routledge. https://doi.org/10.4324/9781315243313-19

Feldman, R. (2013). Techniques and applications for sentiment analysis. *Communications of the ACM*, *56*(4), 82–89. https://doi.org/10.1145/2436256.2436274

Fitz, H., & Chang, F. (2017). Meaningful questions: The acquisition of auxiliary inversion in a connectionist model of sentence production. *Cognition*, *166*, 225–250. https://doi.org/10.1016/j.cognition.2017.05.008

Franco, C. L., & Fugate, J. M. (2020). Emoji face renderings: Exploring the role emoji platform differences have on emotional interpretation. *Journal of Nonverbal Behavior*, *44*(2), 301–328. https://doi.org/10.1007/s10919-019-00330-1

Gohil, S., Vuik, S., & Darzi, A. (2018). Sentiment analysis of health care tweets: Review of the methods used. *JMIR Public Health and Surveillance*, *4*(2), e5789. https://doi.org/10.2196/publichealth.5789

Guerra, P. (2020). Under-connected: Youth subcultures, resistance and sociability in the internet age. In *Hebdige and Subculture in the Twenty-First Century: Through the Subcultural Lens* (pp. 207–230). https://doi.org/10.1007/978-3-030-28475-6_10

Gupta, R. (2019). Data augmentation for low resource sentiment analysis using generative adversarial networks. In *ICASSP 2019–2019 IEEE international conference on acoustics, speech and signal processing (ICASSP)* (pp. 7380–7384). https://doi.org/10.1109/icassp.2019.8682544

Hsu, T. W., Chen, C. C., Huang, H. H., & Chen, H. H. (2021). Semantics-preserved data augmentation for aspect-based sentiment analysis. *EMNLP*, 4417–4422. https://doi.org/10.18653/v1/2021.emnlp-main.362

Hu, E. J., Shen, Y., Wallis, P., Allen-Zhu, Z., Li, Y., Wang, S., Wang, L., & Chen, W. (2021). *Lora: Low-rank adaptation of large language models*. arXiv preprint arXiv:2106.09685. https://arxiv.org/abs/2106.09685

Hussein, D. M. E. D. M. (2018). A survey on sentiment analysis challenges. *Journal of King Saud University-Engineering Sciences*, *30*(4), 330–338. https://doi.org/10.1016/j.jksues.2016.04.002

Jensen, S. Q. (2018). Towards a neo-Birminghamian conception of subculture? History, challenges, and future potentials. *Journal of Youth Studies*, *21*(4), 405–421. https://doi.org/10.1080/13676261.2017.1382684

Jin, W., Zhao, B., Zhang, L., Liu, C., & Yu, H. (2023). Back to common sense: Oxford dictionary descriptive knowledge augmentation for aspect-based sentiment analysis. *Information Processing & Management*, *60*(3), 103260. https://doi.org/10.1016/j.ipm.2022.103260

Kauffmann, E., Peral, J., Gil, D., Ferrández, A., Sellers, R., & Mora, H. (2020). A framework for big data analytics in commercial social networks: A case study on sentiment analysis and fake review detection for marketing decision-making. *Industrial Marketing Management*, *90*, 523–537. https://doi.org/10.1016/j.indmarman.2019.08.003

Liesting, T., Frasincar, F., & Truşcă, M. M. (2021). Data augmentation in a hybrid approach for aspect-based sentiment analysis. In *Proceedings of the 36th annual ACM symposium on applied computing* (pp. 828–835). https://doi.org/10.1145/3412841.3441958

Liu, J., Zhou, Z., Gao, M., Tang, J., & Fan, W. (2023). Aspect sentiment mining of short bullet screen comments from online TV series. *Journal of the Association for Information Science and Technology*, *74*, 1026–1045. https://doi.org/10.1002/asi.24800

Liu, L., Ding, B., Bing, L., Joty, S., Si, L., & Miao, C. (2021). MulDA: A multilingual data augmentation framework for low-resource cross-lingual NER. In *Proceedings of the 59th annual meeting of the association for computational linguistics and the 11th international joint conference on natural language processing* (pp. 5834–5846). https://doi.org/10.18653/v1/2021.acl-long.453

Medhat, W., Hassan, A., & Korashy, H. (2014). Sentiment analysis algorithms and applications: A survey. *Ain Shams Engineering Journal*, *5*(4), 1093–1113. https://doi.org/10.1016/j.asej.2014.04.011

Melo, P. F., Dalip, D. H., Junior, M. M., Gonçalves, M. A., & Benevenuto, F. (2019). 10SENT: A stable sentiment analysis method based on the combination of off-the-shelf approaches. *Journal of the Association for Information Science and Technology*, *70*(3), 242–255. https://doi.org/10.1002/asi.24117

Niu, S., Peng, Y., Li, B., & Wang, X. (2023). A transformed-feature-space data augmentation method for defect segmentation. *Computers in Industry*, *147*, 103860. https://doi.org/10.1016/j.compind.2023.103860

Paltoglou, G. (2016). Sentiment-based event detection in Twitter. *Journal of the Association for Information Science and Technology*, *67*(7), 1576–1587. https://doi.org/10.1002/asi.23465

Pérez Pozo, Á., de la Rosa, J., Ros, S., González-Blanco, E., Hernández, L., & De Sisto, M. (2022). A bridge too far for artificial intelligence?: Automatic classification of stanzas in Spanish poetry. *Journal of the Association for Information Science and Technology*, *73*(2), 258–267. https://doi.org/10.1002/asi.24532

Ren, J., Dong, H., Padmanabhan, B., & Nickerson, J. V. (2021). How does social media sentiment impact mass media sentiment? A study of news in the financial markets. *Journal of the Association for Information Science and Technology*, *72*(9), 1183–1197. https://doi.org/10.1002/asi.24477

Rosenthal, S., Farra, N., & Nakov, P. (2019). *SemEval-2017 task 4: Sentiment analysis in Twitter*. arXiv preprint arXiv:1912.00741. https://arxiv.org/abs/1912.00741

Schmidt, L., & de Kloet, J. (2017). Bricolage: Role of media. In *The international encyclopedia of media effects* (pp. 1–9). Wiley. https://doi.org/10.1002/9781118783764.wbieme0116

Shen, D., Zheng, M., Shen, Y., Qu, Y., & Chen, W. (2020). *A simple but tough-to-beat data augmentation approach for natural language understanding and generation*. arXiv preprint arXiv:2009.13818. https://arxiv.org/abs/2009.13818

Shorten, C., & Khoshgoftaar, T. M. (2019). A survey on image data augmentation for deep learning. *Journal of Big Data*, *6*(1), 1–48. https://doi.org/10.1186/s40537-019-0197-0

Sinha, A., Kedas, S., Kumar, R., & Malo, P. (2022). SEntFiN 1.0: Entity-aware sentiment analysis for financial news. *Journal of the Association for Information Science and Technology*, *73*(9), 1314–1335. https://doi.org/10.1002/asi.24634

Song, M., Feng, Y., & Jing, L. (2023). A survey on recent advances in Keyphrase extraction from pre-trained language models. *Findings of the Association for Computational Linguistics: EACL*, *2023*, 2108–2119. https://aclanthology.org/2023.findings-eacl.161

Sun, Y., & Lee, J. (2020). The relationship between commerce and virtual singer fandom as a subculture: The case of Luo Tianyi.

*International Journal of Art, Culture and Design Technologies*, *4*, 35–42. https://doi.org/10.21742/ijact.2020.4.2.01

Touvron, H., Martin, L., Stone, K., Albert, P., Almahairi, A., Babaei, Y., Bashlykov, N., Batra, S., Bhargava, P., Bhosale, S., Bikel, D., Blecher, L., Ferrer, C. C., Chen, M., Cucurull, G., Esiobu, D., Fernandes, J., Fu, J., Fu, W., … Scialom, T. (2023). *Llama 2: Open foundation and fine-tuned chat models*. arXiv preprint arXiv:2307.09288. https://arxiv.org/abs/2307.09288

Tracz, J., Wójcik, P. I., Jasinska-Kobus, K., Belluzzo, R., Mroczkowski, R., & Gawlik, I. (2020). BERT-based similarity learning for product matching. In *Proceedings of workshop on natural language processing in E-commerce* (pp. 66–75). https://aclanthology.org/2020.ecomnlp-1.7

Verma, S. (2022). Sentiment analysis of public services for smart society: Literature review and future research directions. *Government Information Quarterly*, *39*(3), 101708. https://doi.org/10.1016/j.giq.2022.101708

Wang, Z., Ren, M., Gao, D., & Li, Z. (2023). A Zipf's law-based text generation approach for addressing imbalance in entity extraction. *Journal of Informetrics*, *17*(4), 101453. https://doi.org/10.1016/j.joi.2023.101453

Wang, Z., Zhang, B., & Gao, D. (2022). A novel knowledge graph development for industry design: A case study on indirect coal liquefaction process. *Computers in Industry*, *139*, 103647. https://doi.org/10.1016/j.compind.2022.103647

Wang, Z., Zhang, F., Ren, M., & Gao, D. (2024). A new multifractal-based deep learning model for text mining. *Information Processing & Management*, *61*(1), 103561. https://doi.org/10.1016/j.ipm.2023.103561

Wei, J., & Zou, K. (2019). Eda: Easy data augmentation techniques for boosting performance on text classification tasks. In *Proceedings of the 2019 conference on empirical methods in natural language processing and the 9th international joint conference on natural language processing (EMNLP-IJCNLP)* (pp. 6383–6389). https://doi.org/10.18653/v1/d19-1670

Wong, J., Lee, C., Long, V. K., Wu, D., & Jones, G. M. (2021). "Let's go, baby forklift!": Fandom governance and the political power of cuteness in China. *Social Media+ Society*, *7*(2), 20563051211024960. https://doi.org/10.1177/20563051211024960

Xiang, R., Chersoni, E., Lu, Q., Huang, C. R., Li, W., & Long, Y. (2021). Lexical data augmentation for sentiment analysis. *Journal of the Association for Information Science and Technology*, *72*(11), 1432–1447. https://doi.org/10.1002/asi.24493

Yan, E., Chen, Z., & Li, K. (2020). Authors' status and the perceived quality of their work: Measuring citation sentiment change in nobel articles. *Journal of the Association for Information Science and Technology*, *71*(3), 314–324. https://doi.org/10.1002/asi.24237

Yang, Y., Malaviya, C., Fernandez, J., Swayamdipta, S., Bras, R. L., Wang, J. P., Bhagavatula, C., Choi, Y., & Downey, D. (2020). *Generative data augmentation for commonsense reasoning*. arXiv preprint arXiv:2004.11546. https://arxiv.org/abs/2004.11546

Yildirim, G. (2022). A novel grid-based many-objective swarm intelligence approach for sentiment analysis in social media. *Neurocomputing*, *503*, 173–188. https://doi.org/10.1016/j.neucom.2022.06.092

Yule, G. (2022). *The study of language*. Cambridge University Press.

Zea, Q., & Heekyoung, J. (2019). Learning and sharing creative skills with short videos: A case study of user behavior in tiktok and bilibili. In *Int. Assoc. Soc. Des. Res. Conf* (No. 10, pp. 25–50). https://iasdr2019.org/uploads/files/Proceedings/le-f-1209-Zho-Q.pdf

Zhang, L., Wang, S., & Liu, B. (2018). Deep learning for sentiment analysis: A survey. *Wiley Interdisciplinary Reviews: Data Mining and Knowledge Discovery*, *8*(4), e1253. https://doi.org/10.1002/widm.1253

Zhang, X., Zhao, J., & LeCun, Y. (2015). Character-level convolutional networks for text classification. In *Advances in Neural Information Processing Systems* (pp. 649–657). https://arxiv.org/abs/1502.01710